\documentclass{llncs}
\usepackage[utf8]{inputenc}
\usepackage{xcolor}
\usepackage{comment}
\usepackage{graphicx}
\usepackage{hyperref}
\usepackage{xspace}
\usepackage{enumitem}
\setlist{leftmargin=*}
\newcommand{\shiftup}[0]{\vspace{-0.15cm}}
\newcommand{\ttitle}[0]{\texttt{TipsC}}
\newcommand{\stitle}[0]{TipsC}
\title{\stitle{}: Tips and Corrections for programming MOOCs}
\author{Saksham Sharma \and Pallav Agarwal \and Parv Mor \and Amey Karkare}
\institute{Indian Institute of Technology, Kanpur\\
\email{\{sakshams, pallavag, parv, karkare\}@cse.iitk.ac.in}}

\begin{document}

\maketitle

\begin{abstract}
With the widespread adoption of MOOCs in academic institutions, it has become imperative to come up with better techniques to solve the tutoring and grading problems posed by programming courses. Programming being the new `writing’, it becomes a challenge to ensure that a large section of the society is exposed to programming. Due to the gradient in learning abilities of students, the course instructor must ensure that everyone can cope up with the material, and receive adequate help in completing assignments while learning along the way.

We introduce \ttitle{} for this task. By analyzing a large number of correct submissions, \ttitle{} can search for correct codes resembling a given incorrect solution. Without revealing the actual code, \ttitle{} then suggests changes in the incorrect code to help the student fix logical runtime errors. In addition, this also serves as a cluster visualization tool for the instructor, revealing different patterns in user submissions.

We evaluated the effectiveness of \ttitle{}'s clustering algorithm on data collected from previous offerings of an introductory programming course conducted at IIT Kanpur where the grades were given by human TAs. The results show the weighted average variance of marks for clusters when similar submissions are grouped together is 47\% less compared to the case when all programs are grouped together.

\end{abstract}


\section{Introduction}
With Massively Open Online Courses (MOOCs) being widely adopted among academic institutions and online platforms alike, the number of students studying programming through such courses has sky-rocketed. In contrast, the availability of personalized help through Teaching Assistants (TAs) can not scale accordingly due to human limitations.

The challenge in this scenario is two-fold. Firstly, TAs have to manually grade a large number of incorrect submissions for partial grades, and this process is prone to a large bias and variance as we discovered through collected data~\cite{gradeit}. Secondly, helping students stuck at a problem (by providing relevant tips and suggestions) is simply not tractable for MOOCs due to the scale involved.

We introduce \ttitle{}, a tool to parse, analyze, and cluster programming MOOC submissions, in order to tackle the above challenges.

\subsection*{Motivational Example}
\begin{figure}[t!]
\begin{minipage}[c]{0.59\textwidth}
\begin{verbatim}
  int main() {
    int n; float f;
    scanf("%d %f", &n, &f);
    if (n%5==0 && (n+.5)<=f && f<=2000)
      printf("%0.2f", f-n-.5);
    else
      printf("%0.2f", f);
  }
\end{verbatim}
\end{minipage}
\begin{minipage}[c]{0.39\textwidth}
\begin{verbatim}
int main() {
  float y; int x;
  scanf("%d %f", &x, &y);
  if ((x%5==0) && (x+0.5)<=y)
    printf("%0.2f", y-x-.50);
  else
    printf("%0.2f", y);
}
\end{verbatim}
\end{minipage}
\caption{Two programs with a small logical difference. \newline
\hspace*{1.1cm}\textbf{Source:} codechef.com/problems/HS08TEST.
\label{fig:motiv}}
\shiftup
\end{figure}

Consider the two submissions in an offering of a programming
course as shown in Fig~\ref{fig:motiv}. It would be ideal if
the author of the second program could be informed about the
missing bound check, once the author has tried enough. Using a
large number of submissions available for each problem, \ttitle{}
finds programs similar to the incorrect one, and suggests
granular changes.

The algorithm to find similar submissions allows \ttitle{} to
cluster the programs together. This aids in manual analysis,
allowing instructors and TAs to obtain a bird's-eye view of the spectrum of solutions received from the students and also helps
in creating customized feedback and grading rubrics for
individual clusters.

\subsection*{Our Contributions}
\ttitle{} solves a very practical use-case, which has not been explored enough in the past. The contributions of this paper are:
\begin{enumerate}
\item A technique to normalize C program Abstract Syntax Trees (ASTs) to a linear representation, with an intention to make them amenable to similarity analysis. This includes various approximations and domain-specific heuristics.
\item An edit distance metric for normalized programs, which is a set of specialized modifications upon Levenshtein distance for two lists.
\item We demonstrate the effectiveness of our distance metric by clustering similar programs (distance within a certain threshold) together and comparing the variance in marks awarded by TAs within a cluster.
\item An open source tool, \ttitle{}, which implements the above ideas in the context of a MOOC teaching the C language.

\end{enumerate}

\section{\stitle{}}
\label{sec:workflow}
\ttitle{}, implemented in Scala, can be plugged into a MOOC to realize the contributions mentioned above. The primary motivation behind the working of the software is the fact that most introductory programming assignments have a finite number of solution variants, with minor variations in between them. We assume that a student's solution would often resemble some previously existing solution(s). \ttitle{} attempts to find programs which are similar to a user's attempt. It can then suggest changes to the user's program, which would involve fixing small logical runtime errors.

The software offers a \textit{command line interface} (CLI), as well as a \textit{web interface}, with an intersecting set of functionality available. It is intended to be run as a web service behind the scenes, in a MOOC. The CLI's purpose is to allow experimenting with program similarity metrics, and for producing illustrative figures to aid in visualization of the data.

\subsection*{Workflow}
\begin{enumerate}
\item \ttitle{} accepts C language programs as input, which it parses using an inbuilt parser. Relevant features are extracted from the parsed result, and the program is normalized and converted to a linear representation (Sections \ref{subsec:proglinear} and \ref{subsec:prognormal}). All this information is serialized and stored in a database.
\item On every valid program insert request, edit distances (Section \ref{subsec:editdist}) between that program and all existing submissions (for that particular question) are computed, using the method described in Section \ref{subsec:comparing}.
\item Periodically, the distance matrix is consumed by a script which creates clusters out of the provided programs (Section \ref{subsec:cluster}). Such clusters are formed for each active problem in the MOOC. The number of clusters is ensured to be sub-polynomial in $n$.
\item \ttitle{} provides an endpoint which accepts a valid C program, which is then linearized and normalized. Edit distance is computed between the input program and representative elements from our clusters, and then to each element of the closest clusters, which allows us to select the closest programs to our input program.
\item The edit distance metric discussed in Section \ref{subsec:editdist} also returns a `patch' to convert the normalized programs into each other. Thus, knowing the closest programs allows \ttitle{} to provide personalized tips for the problem to the user. Of course, such tips must be filtered appropriately in order to prevent leaking solutions. This is discussed in section \ref{sec:usage}.
\end{enumerate}

\section{Algorithms}
In this section, we provide details on the various algorithms used by the workflow in Section~\ref{sec:workflow}. The programs submitted by the students, can not be compared as is. To use our variation of the Levenshtein distance algorithm, we first normalize all the programs (into a linear form) by a series of transformations on the obtained AST of the program. After this, our metric for comparing programs, and our clustering technique, together allow \ttitle{} to efficiently suggest changes in programs, as discussed above.

\subsection{Program normalization}
\label{subsec:prognormal}
We now describe various stages of program normalization.
\vspace{-0.2cm}
\subsubsection{Linearization}
\label{subsec:proglinear}
The program is first converted to a linear representation, rather than one with nested constructs. This is done to aid in the comparison step. For example, an if-else construct would be converted to:\\
\texttt{IF(condition)~BLOCK\_START...BLOCK\_END~ELSE~BLOCK\_START...BLOCK\_END}.\\ Ideally, each statement would translate to one or more tokens in the linear form.

\vspace{-0.1cm}
\subsubsection{Construct Normalization}
Since there are multiple loop constructs, they are replaced with the closest approximation of a single LOOP construct. For instance, a for loop may be split into an assignment, a while loop, and a update operation at the end of the while block. \texttt{"for(expr1; cond1; expr2) BLOCK"} becomes \texttt{"expr1; loop(cond2); (BLOCK + expr2);"}) This allows for a better similarity measure between two programs. Similarly, other semantically similar constructs could be handled (for instance, renaming \texttt{var++} to \texttt{var=var+1}).

\vspace{-0.1cm}
\subsubsection{Expression Linearization}
Just like the flow graph, we also need a linear representation of the expressions. For our purposes, we use the postfix notation for the expression linearization.

\vspace{-0.1cm}
\subsubsection{Expression normalization}
Since different students use different variable names, we rename the variables in the expression to generic names, based on their order of use \textit{within that expression}.
For instance, the expression $(a + b / a)$ after normalization would become: $(var_1 + var_2 / var_1)$. Or, in postfix: $var_1\ var_2\ var_1\ /\ +$

\vspace{-0.1cm}
\subsection{Edit distance}
\label{subsec:editdist}
After the linearization of the program, we use a variation of the Levenshtein edit distance algorithm to find the similarity between two programs. The edit distance algorithm is run on the linear representation of the programs, with each logical line of code considered as one token. We assign a constant value ($W_{ad}$) for additions and deletions, and a maximum value for ($W_r = W_{ad}/2$) for replacements of most constructs.

Since it is not enough to compare just the equality of two statements/expressions, a granular edit cost is computed when the tokens being compared are both expressions. The edit cost inside the expressions is computed in exactly the same manner as the rest of the program (Levenshtein). Later, the edit cost when comparing expressions is normalized by the size of the expression and finally scaled to $W_r$, where an edit cost of $W_r$ implies completely different expressions.

A simplified pseudocode of the algorithm is given in Fig~\ref{fig:editmodule}.

\begin{figure}[t!]
\begin{verbatim}
func EditModule(list1, list2: List[Item]):
  let head1, head2 = head(list1), head(list2)

  if head1 == head2:
    return EditModule(tail(list1), tail(list2))

  normalized_dist = 10
  if type(head1) == Expr && type(head2) == Expr:
    partial_distance = EditModule(head1, head2)
    normalized_dist = normalize(partial_distance, head1, head2)

  distance1 = 20 + EditModule(list1, tail(list2))
  distance2 = 20 + EditModule(tail(list1), list2)
  distance3 = normalized_dist + EditModule(tail(list1), tail(list2))

  return min(distance1, distance2, distance3)
\end{verbatim}
\vspace{-0.2cm}
\caption{Pseudo-code for edit distance algorithm\label{fig:editmodule}}
\end{figure}

\begin{figure}[t!]
\begin{minipage}[c]{0.59\textwidth}
\begin{verbatim}
int func1() {
  int r = func2();
  return func4(r);
}

int func2() { return func3(); }

int func3() { return 2; }

int func4(int r) { return r-2; }

int main() { return func1(); }
\end{verbatim}
\end{minipage}
\begin{minipage}[c]{0.39\textwidth}
\includegraphics[width=1.4in]{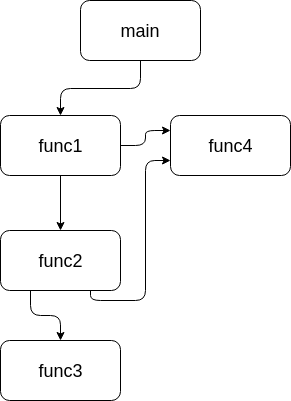}
\end{minipage}
\caption{Depth-First traversal: \texttt{main $\rightarrow$ func1  $\rightarrow$ func2  $\rightarrow$ func3  $\rightarrow$ func4}}
\label{fig:dfs}
\vspace{-0.5cm}
\end{figure}

In addition to the above, we intentionally penalize addition or deletion of \texttt{BLOCK\_OPEN} and \texttt{BLOCK\_CLOSE} constructs with thrice the usual penalty, so that the matching algorithm tries to align blocks with each other, wherever possible. This modification should make the suggestions much more meaningful since now loops and conditionals would be matched to corresponding constructs with a higher likelihood. This is because a higher penalty for block anchors incentivizes matching-up blocks between the codes.

\subsection{Comparing Two Programs}
\label{subsec:comparing}
\ttitle{} uses a modified edit distance algorithm (Section~\ref{subsec:editdist})
to compare normalized programs. This, alone, is not enough to capture the variations seen in programs. This is because naively detecting homomorphic reordering of statements (reordering content across functions or changing function order) would make the comparison algorithm very inefficient. We solve this issue by comparing each function separately.


Each function is represented as a list of tokens, and a program is represented as a list of function representations. In addition, we do a Depth-First-Traversal on expressions in the `main' function, to reorder all the functions in the program in the order they are first called. Such a traversal also removes any unused functions from the analysis.
\noindent It is easy to imagine how introductory programming assignments could involve multiple functions which call each other, and how ordering them in first-use-order would compensate for the many possible ways students could choose to order their functions. An example of such a traversal is shown in Fig~\ref{fig:dfs}.\\
\begin{figure}[t!]
\begin{minipage}[c]{0.54\textwidth}
\begin{verbatim}
            if (0)
              helper1();
            else
              helper2();
\end{verbatim}
\end{minipage}
\begin{minipage}[c]{0.44\textwidth}
\begin{verbatim}
    if (1)
      helper2();
    else
      helper1();
\end{verbatim}
\end{minipage}
\caption{Example where first-use-based ordering does not work}
\label{fig:firstuse}
\vspace{-0.1cm}
\end{figure}
\noindent There are still two scenarios remaining. A user could choose to write a singleton `main' function, which calls another helper function for the main logic. Another scenario where the first-use-ordering would be incorrect is shown in Fig~\ref{fig:firstuse}.\\
\begin{figure}[t!]
\begin{verbatim}
func PairUpFunctions(fxns1, fxns2: List[Function]):
  let result = []

  for permu in permutations(fxns2):
    pairs    = zip/pair up functions in fxns1, permu in order

    leftOut1 = remaining (unpaired) functions in fxns1 if any
    leftOut2 = remaining (unpaired) functions in permu if any
    leftOutLines = sum of length of tokens in leftOut functions

    leftOutPenalty  = leftOutLines * scalingFactor1
    orderingPenalty = (fraction of functions in permu which are not
                       in their use-order position) * scalingFactor2
    add (pairs, leftOutPenalty + orderingPenalty) to result
  return result
\end{verbatim}
\vspace{-0.3cm}
\caption{Pseudo-code for function pairing algorithm\label{fig:algo}}
\vspace{-0.7cm}
\end{figure}
\noindent Both these scenarios are taken care of by our algorithm, shown in Fig~\ref{fig:algo}.
The algorithm considers all permutations of the functions, assigning appropriate penalties in case the ordering is different or if some functions could not be matched.
Note that this pairing algorithm is exponential in the number of functions present in the programs. Yet, it is practical since we do not expect more than a handful of functions in an introductory course. Even in cases where a large number of functions are expected in a program, the implementation of the algorithm can be forced to time out after a threshold.


\subsection{Clustering programs}
\label{subsec:cluster}
Since the distances obtained from the algorithm in Section~\ref{subsec:editdist} do not follow triangle inequality, we cannot map the programs onto a vector space that accurately captures their distance matrix. Instead we use hierarchical/agglomerative clustering~\cite{clustering} which does not require assertion of the inequality. We perform clustering using the following four different linking criteria, which are: \textit{single}, \textit{complete}, \textit{average} and \textit{weighted}.

For each of these methods, we calculate cophenetic correlation coefficient~\cite{coeff} and select the linkage method which maximizes the coefficient. Based on this method we generate the hierarchical tree and create clusters using the algorithm in Fig~\ref{fig:pruning}.
We define \verb{thresholdCount{ to be $\lfloor{\sqrt{n}}\rfloor{}$, where $n$ is the number of programs and \verb{thresholdDist{ is the distance only below which creation of clusters will be allowed to filter out the outliers.

\begin{figure}[t!]
\begin{verbatim}
    func DoCluster(rootNode: Node):
        if rootNode is None:
            return
        else if rootNode.dist > thresholdDist
                || rootNode.count >= 2 * thresholdCount:
            DoCluster(rootNode.left) && DoCluster(rootNode.right)
        else if rootNode.left is None || rootNode.right is None:
            DoCluster(rootNode.left) && DoCluster(rootNode.right)
        else if rootNode.count <= thresholdCount:
            MakeNewCluster(rootNode)
        else:
            DoCluster(rootNode.left) && DoCluster(rootNode.right)
\end{verbatim}
\caption{Pseudo-code for creating hierarchical clusters}
\label{fig:pruning}
\vspace{-0.3cm}
\end{figure}

\subsection{Finding representative elements of clusters}
\label{subsec:repcluster}
In \ttitle{}, the concept of representative elements is introduced to avoid computing distance with all the programs in the database with the input program. 
{Clustering} the tree using algorithm in Fig~\ref{fig:pruning} ensures that we have $\Omega(\sqrt{n})$ clusters with each cluster having $\mathcal{O}(\sqrt{n})$ elements. As a result to find the nearest program we compare the input program first with representative elements of each of the $\Omega(\sqrt{n})$ clusters. Then filter out on the second level with every element of the starting best clusters.

\noindent To find out the representative element of a cluster we the take the program from it which gives the least root mean square error with all elements of the cluster.

\section{Usage in a MOOC environment}
\label{sec:usage}
\ttitle{} approach is particularly suited for a programming MOOC.
MOOCs usually allow a few days of time for solution submission. Students may start attempting the problem early on, which would ensure that a large number of correct solutions populate the database in some time. Students having difficulty with the problem may be unable to submit a correct solution in time, since they may find themselves stuck on minor errors. When a particular amount of time has passed, the instructor may choose to activate \ttitle{}, which would allow slower students to get hints using the already processed correct solutions. The instructor may choose to penalize all submissions after that particular time, which is a standard practice in many university courses even without \ttitle{}.

The above approach would ensure that students do not stay stuck at a particular problem and give up, but rather can take advantage of automated and personalized hints to proceed further in their assignments, thus hopefully lowering the drop-out rate. Also, since this algorithm can potentially scale to very large courses, this would also reduce the number of TAs required for MOOCs.

There is a fine line between helping and spoon-feeding, and \ttitle{} tries its best not to cross it. The suggestions provided by \ttitle{} are not plug-and-play, but are rather hints. The difference between programs is on a processed version of the program and does not contain exact variable names or even syntax. For example, a user may notice that there is a missing conditional check, which may bring their attention to that part of the code, but they would not see an already complete solution. They would also not be shown large differences, thus preventing a leak of program logic or structure.

\section{Implementation and Scaling}
\ttitle{} is an open source software, released under the Apache 2.0 License, on GitHub.\footnote{The source code is available at \url{https://github.com/HexFlow/tipsy}}. The major implementation is in Scala, using the Akka framework for the backend. The clustering algorithm used is implemented in SciPy and is run using Python scripts. A web playground has been deployed publicly, which allows users to see the algorithm live in action on different problems.\footnote{The web playground for \ttitle{} is deployed at \url{http://tipsy.hexflow.in}}

The performance was tested on a Linux desktop running OpenJDK8, with 16 GB RAM, and an Intel(R) Core(TM) i7-4470 CPU @ 3.40GHz with 4 cores.

\subsubsection{Program Addition}
This entails a simple addition into the database, fetching the distance matrix for the background distance update job, and starting the background jobs. Thus, there is no noticeable time required for this step.

\subsubsection{Updating the Distance Matrix}
For every new program addition, distances with existing programs must be computed. Requests are handled sequentially to avoid race conditions. If the database has $n$ programs, then this part of the algorithm takes $O(n)$ time to compute the distance for each of those programs.

\begin{figure}[t!]
  \centering
  \begin{tabular}{@{}p{.5\textwidth}@{}p{.5\textwidth}@{}}
    \includegraphics[width=.50\textwidth]{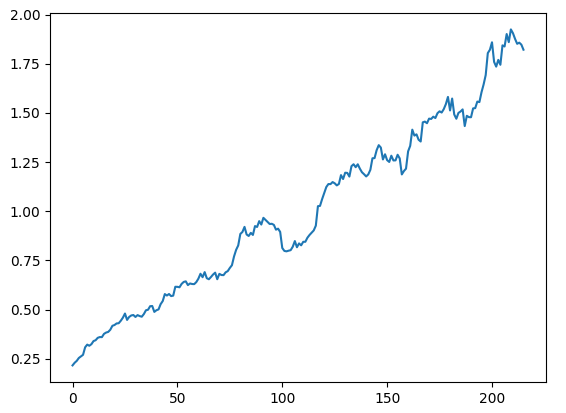}
    &
    \includegraphics[width=.495\textwidth]{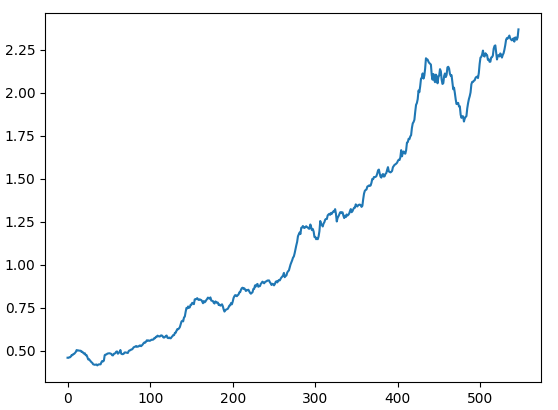}
    \\
    \multicolumn{1}{c}{(a) Distance Matrix Update }
    &
    \multicolumn{1}{c}{(b) Cluster Update }
  \end{tabular}
    \caption{Performance on an average of 50 lines of code per program. The x-axis denotes the number of already submitted programs. Y-axis denotes the time taken in seconds to update the data structures on adding a new program.}
    \label{fig:distupdate}
    \label{fig:clustupdate}

\end{figure}

This step is only run for correct submissions. This is bounded by the number of students enrolled in the course, and thus would be quite infrequent. Yet, this step is amenable to parallelization. This makes its performance very practical for MOOCs, as is shown in Fig~\ref{fig:distupdate}(a)

\subsubsection{Updating Clusters}
This part of the algorithm is not run frequently, since it only provides more programs to be matched against, for correction fetching. This part is fast enough on practical database sizes, but it scales as $O(n^2)$. This is not a major overhead, as is shown in Fig~\ref{fig:clustupdate}(b)

\subsubsection{Providing Corrections}
Corrections are served by matching against representative elements of each cluster, and later by comparing against all elements of 4 best clusters. Since the number of programs in a cluster is bounded (See Section~\ref{subsec:cluster}), and the number of clusters is expected to be small. The worst case is still $O(n)$ where all submissions are very far away from each other. This scenario is expected to be very rare for usual introductory problems of programming. On real data, from the Introduction to Programming course at IIT Kanpur, we see 30-40 clusters among 100 submissions for each problem, around 20-30 of which are singleton clusters of outliers. These numbers are from Week 3 and Week 4 problems, which contain simple recursion problems, multiple nesting loops, and conditionals, among other things.

\noindent In the scenario described above, requests for fetching corrections for non-trivial codes (35-45 lines of code) take 0.6-0.8 seconds end-to-end. This is expected to scale slowly with the number of programs in the database and thus is quite feasible for MOOCs. The major overhead is due to the comparison with representative elements of each cluster, and this task can be made faster with more parallelism.

\section{Experiments}
\begin{figure}[t!]
\makebox[\textwidth][c]{\includegraphics[width=1.0\textwidth]{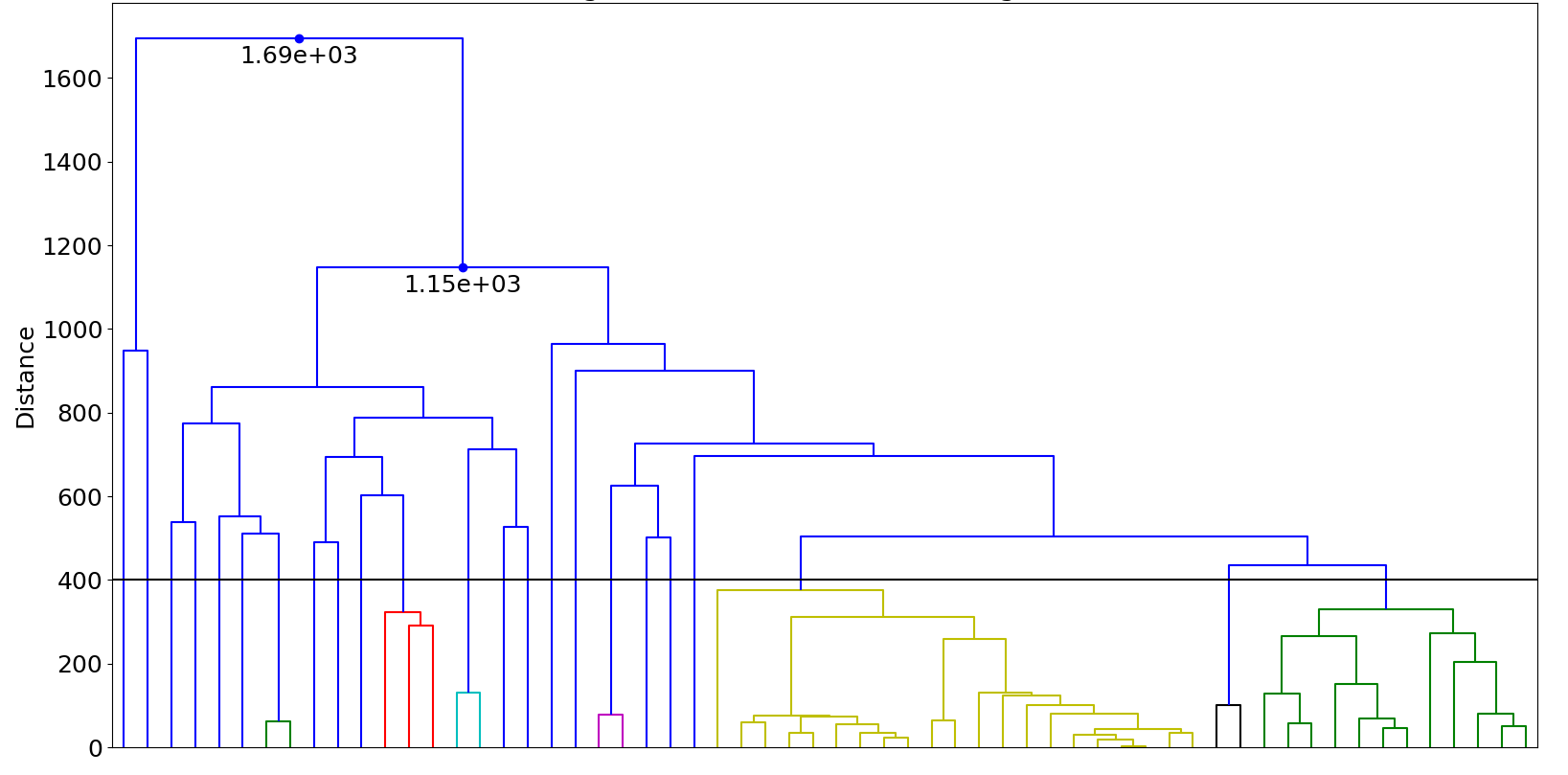}}
\caption{Dendrogram from the clustering algorithm. X-axis represents unique programs.}
\label{fig:dendro2}
\vspace{-0.5cm}
\end{figure}

\begin{figure}[t!]
\makebox[\textwidth][c]{\includegraphics[width=0.5\textwidth]{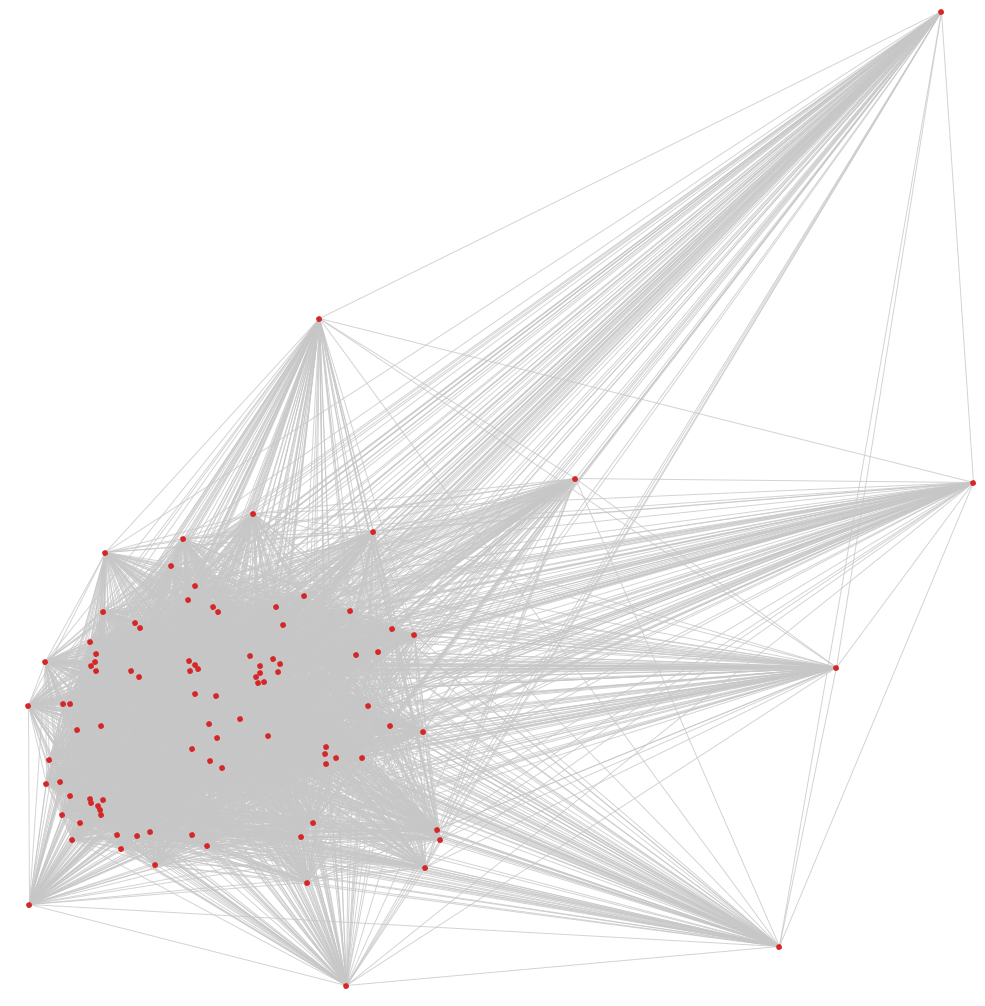}}
\caption{Force-graph visualization of submissions to a problem}
\label{fig:force}
\end{figure}

\ttitle{} was run on data from previous iterations of the Introduction to Programming course at the authors' institute. Some observed trends are described in Fig~\ref{fig:dendro2}, Fig~\ref{fig:force} and Table~\ref{table:vari}.

\noindent Clustering followed by manual inspection of the clusters revealed some inconsistency in manual grading by TAs (among partially correct submissions). We inspected the variance of marks in each non-trivial cluster, which was often low (since many submissions got full marks). On inspecting clusters with high marks variance, we noticed programs with minor differences, but with disparate marks. Clustering on 85 submissions to a fairly advanced problem containing loops, conditionals and arrays, yielded a small number of interrelated clusters, as verified by manual inspection (Fig~\ref{fig:dendro2})

To compare the effectiveness of the clustering by \ttitle{}, we computed the variance of marks in each cluster for several problems as shown in Table~\ref{table:vari} (Problem ID is a unique ID given to each assignment for the course). The results show that variance within a cluster is much less (47\% less on average) than when all the submissions are considered together. This suggests that \ttitle{} is indeed able to group similar programs together, a fact that can help in effective grading by assigning similar programs to the same TA.

\ttitle{} also generates force-layout based images for visualization purposes of the submissions. Figure~\ref{fig:force} shows an example, when run on 100 submissions to a problem requiring solutions with 4-5 nested loops and conditionals. It also helps pinpoint outliers, which can be inspected manually by the instructor later.

\begin{table}[ht!]
\centering
\def\arraystretch{1.1}
\setlength{\tabcolsep}{.25em}
\begin{tabular}{|c|c|c|c|}
\hline
Problem ID & \# submissions & Variance (overall) & Average cluster variance \\ \hline
Lab3-1633  & 84             & 1.54                      & 0.78                                 \\ \hline
Lab4-1822  & 68             & 2.15                      & 0.70                                 \\ \hline
Lab6-2012  & 64             & 3.33                      & 1.97                                 \\ \hline
Lab8-2289  & 68             & 1.92                      & 1.30                                 \\ \hline
Exam1-1938 & 69             & 6.93                      & 3.74                                 \\ \hline
\end{tabular}
\vspace{0.2cm}
\caption{Comparison of variance of marks with and without clustering\label{table:vari}}
\vspace{-0.3cm}
\end{table}

\section{Related Work}

Automated repair and feedback generation for introductory programming courses is attracting plenty of attention in recent years due to plethora of online courses available on MOOC platforms, and due to growing number of students in traditional classrooms. The repair or feedback can be produced at two different stages for student programs: (a) when a student is struggling to fix compile time errors, and (b) when the program is running, but the student is unable to match the behavior expected by the instructor (typically specified through unit tests).
\newcommand{\myalgo}{TRACER\xspace}

\myalgo~\cite{umair18icse} learns fixes for compile time errors from existing code submissions (possibly for a different problem statement) and performs targeted repair. The learning is performed by comparing an erroneous version and a later error-free version of the same program. Other methods~\cite{aditya,rishabh,armando} learn repairs by observing a large corpus of correct programs to learn possible correct sequences of tokens.
\emph{HelpMeOut}~\cite{hartmann2010would} generates feedback for compilation error by maintaining a database of errors encountered by other students. In case of a compiler error,  it provides both the erroneous line of the other student, as well as the modified line which resulted in successful compilation as a hint. {\em GradeIT}~\cite{gradeit} uses simple rewrite rules to repair common compilation errors. Their study show that even these simple  repairs can be effective for feedback generation and automated grading of assignments.

The Software Engineering community has developed a number of {\em Automated program repair} (APR)\footnote{\url{http://program-repair.org/}} tools (GenProg~\cite{genprog}, AE~\cite{Weimer-ASE13}, Angelix~\cite{angelix}, and Prophet~\cite{prophet} to name a few) that automatically fix software bugs. These tools have been shown to fix the bugs of large real-world software effectively. However, a recent study~\cite{fse2017prutor} has concluded that these repairs are not directly suitable to be used as hints for novice students programmers. On the other hand, they can be an effective aid for improving grading by teaching assistants (typically student programmers with few years of experience).

REFAZER~\cite{refazer} uses program synthesis, particularly programming-by-example technique,  to synthesize syntactic program transformations to fix logical errors in the program. It uses a corpus of code edits made by students to fix incorrect programs to generate transformers that can be used later on similar incorrect programs submitted by other students. Other approaches~\cite{RiversK14,RiversK13,RiversK17,priceDB16} use reference solution(s) from the instructor and correct/incorrect programs from other students to construct a solution space containing the different states (correct and incorrect) that students have created. Then, a path from an incorrect state to some nearest correct state is used to generate hints.

Another common approach to feedback generation is clustering where students submissions having similar features are grouped together in clusters. The clusters are created either by using a fixed set of rules or by using machine learning techniques~\cite{codeweb,embedding,gross} or by using techniques based on program analysis~\cite{radicek2,radicek,shalini,overcode,head}. The feedback is typically generated manually for a representative program in each cluster, and it is customized to other members of the cluster automatically. Our tool \ttitle{} belongs to the category of rule-based clustering tools. However, it differs significantly from others in the use of linearized ASTs that are amenable to efficient distance computation between two programs. Comparisons with TA marks show that these distances can be used as a measure of correctness for these programs.

\section{Conclusion and Future Work}
In this paper, we have described \ttitle{} in the context of programming and logical error corrections. It is a scalable system, which can be plugged into any existing MOOC, to allow aiding students who are having difficulty in the course without any manual intervention. The system requires a very reasonable number of submissions to become practical, and can easily be modified to handle many other programming languages as well. It works in a fully automated manner and does not require any special effort to accommodate different problems.

We plan to deploy \ttitle{} in an offering of the Introduction to Programming course at our institution, and conduct user surveys to evaluate its usefulness. We believe that \ttitle{} rules can be easily mapped to create helpful feedback messages and rubrics to grade programs with minimum human intervention. Also, some lightweight semantic analysis, and inlining of non-recursive functions, can be used to improve the similarity matrix. We plan to implement these in a future iteration of the software.
Use of \ttitle{} as a plagiarism detector can also be explored.

\vspace{1cm}

\bibliographystyle{splncs_srt}
\bibliography{main}

\end{document}